%% file: main.tex
\documentclass[sigconf]{acmart}
\setcitestyle{numbers,sort&compress}

\usepackage{latexsym}
\usepackage{amsmath}
\usepackage{amsthm}
\usepackage{booktabs}
\usepackage{enumitem}
\usepackage{graphicx}
\usepackage{color}
\usepackage{cleveref}
\usepackage{xcolor}
\usepackage{multirow}
\usepackage{cleveref}
\usepackage{float}
\usepackage{diagbox}
\usepackage{subcaption}
\usepackage{placeins} 
\usepackage{url}

\crefname{figure}{Fig.}{Figs.}
\Crefname{figure}{Fig.}{Figs.}
\crefname{table}{Tab.}{Tabs.}
\Crefname{table}{Tab.}{Tabs.}

\AtBeginDocument{%
  }

\acmConference[Conference acronym 'XX]{Make sure to enter the correct
  conference title from your rights confirmation email}{June 03--05,
  2018}{Woodstock, NY}

\begin{document}

\title{Transferable Physical-World Adversarial Patches Against Pedestrian Detection Models}


\author{Shihui Yan}
\email{yanshihui@hust.edu.cn}
\affiliation{%
  \institution{Huazhong University of Science and Technology}
  \country{China}
}
\author{Ziqi Zhou}
\email{zhouziqi@hust.edu.cn}
\affiliation{%
  \institution{Huazhong University of Science and Technology}
  \country{China}
}
\author{Yufei Song}
\email{yufei17@hust.edu.cn}
\affiliation{%
  \institution{Huazhong University of Science and Technology}
  \country{China}
}
\author{Yifan Hu}
\email{hyf1009@hust.edu.cn}
\affiliation{%
  \institution{Huazhong University of Science and Technology}
  \country{China}
}
\author{Minghui Li}
\email{minghuili@hust.edu.cn}
\affiliation{%
  \institution{Huazhong University of Science and Technology}
  \country{China}
}
\author{Shengshan Hu}
\authornote{Shengshan Hu is the corresponding author.}
\email{hushengshan@hust.edu.cn}
\affiliation{%
  \institution{Huazhong University of Science and Technology}
  \country{China}
}

\renewcommand{\shortauthors}{}

\input{Section/0-abstract}



\maketitle

\input{Section/1-introduction}

\input{Section/2-relatedwork}
\input{Section/3-methodology}

\input{Section/4-experiments}
\input{Section/5-conclusion}

\bibliographystyle{unsrt}
\bibliography{main}

\end{document}

%% file: Section/0-abstract.tex
\begin{abstract}
Physical adversarial patch attacks critically threaten pedestrian detection, causing surveillance and autonomous driving systems to miss pedestrians and creating severe safety risks.
Despite their effectiveness in controlled settings, existing physical attacks face two major limitations in practice: 
they lack systematic disruption of the multi-stage decision pipeline, enabling residual modules to offset perturbations, and they fail to model complex physical variations, leading to poor robustness.
To overcome these limitations, we propose a novel pedestrian adversarial patch generation method that combines multi-stage collaborative attacks with robustness enhancement under physical diversity, called TriPatch.
Specifically, we design a triplet loss consisting of detection confidence suppression, bounding-box offset amplification, and non-maximum suppression (NMS) disruption, which jointly act across different stages of the detection pipeline.  
In addition, we introduce an appearance consistency loss to constrain the color distribution of the patch, thereby improving its adaptability under diverse imaging conditions, and incorporate data augmentation to further enhance robustness against complex physical perturbations.
Extensive experiments demonstrate that TriPatch achieves a higher attack success rate across multiple detector models compared to existing approaches.


\end{abstract}


%% file: Section/1-introduction.tex
\section{Introduction}\label{sec:inroduction}
Deep neural networks~\cite{xue2025towards,pan2026ufvideo,wu2025tattoo} achieve remarkable progress in object detection and are widely deployed in security surveillance and autonomous driving. Modern detectors provide accurate localization and classification under complex real-world scenes and serve as fundamental components in intelligent transportation and smart monitoring systems. 
However, despite their strong performance, these systems remain exposed to numerous security threats~\cite{zhou2025darkhash, badhash,wan2025mars,zhang2024detector,zhang2025test,wang2024trojanrobot,li2025detecting,wang2024eclipse,yu2025spa,wang2024unlearnable}, particularly exhibiting high vulnerability to adversarial attacks~\cite{carlini2016towards,wang2025advedm,wang2025breaking,song2025segment}. Even small, carefully crafted perturbations can significantly degrade detection performance and expose critical security risks in safety-sensitive applications.

\begin{figure}[!t]
    \centering
\includegraphics[scale=0.38]{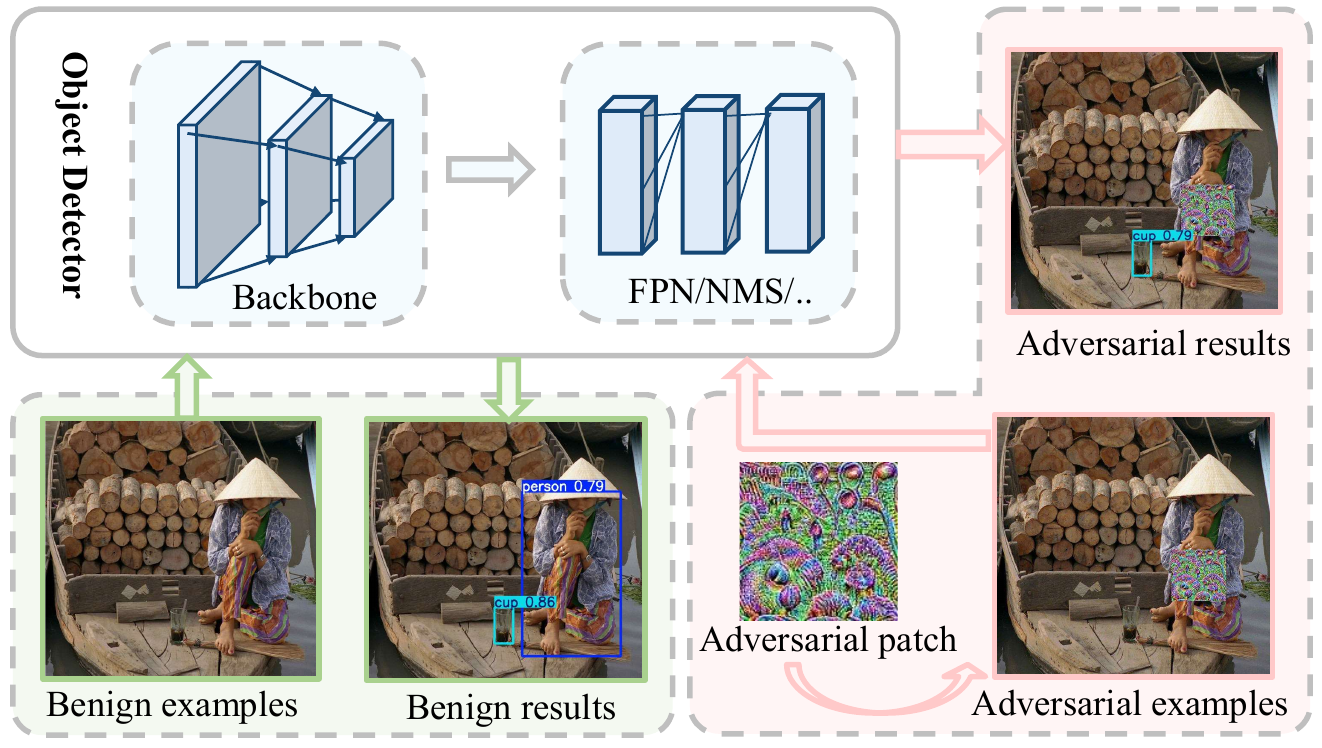}
    \caption{Overview of adversarial examples on object detectors.
    }
    \label{fig:demo}
       \vspace{-0.4cm}
\end{figure}

The security vulnerabilities of object detection systems pose significant challenges to multimedia retrieval and analysis applications. In security surveillance and intelligent monitoring systems, adversarial attacks can compromise the integrity of multimedia content analysis, leading to incorrect indexing, failed event detection, and unreliable semantic understanding. When detection models fail to accurately identify pedestrians due to adversarial patches, downstream multimedia applications such as video summarization, content-based retrieval, and automated annotation systems inherit these errors, potentially causing cascading failures in large-scale multimedia databases. Moreover, the robustness of multimedia machine learning systems becomes critical when deploying deep neural networks for real-world multimedia analysis tasks. Understanding and addressing these adversarial vulnerabilities is essential for developing trustworthy multimedia retrieval systems that can reliably process and analyze visual content in security-sensitive environments.

Adversarial examples are input samples with subtle perturbations that appear benign to humans but mislead models into producing incorrect predictions. In particular, adversaries attach adversarial patches~\cite{guesmi2024dap,hu2023physically} to the human body, misleading detection models without altering the overall image content and, in some cases, completely bypassing detection. Such attacks are especially concerning in pedestrian detection, where failures directly threaten traffic safety and public security.


Adversarial attacks are classified as digital~\cite{li2024transferable,song2025seg,song2026erosion,wangadvedm} or physical~\cite{thys2019fooling,wang2021dual,xu2020adversarial}. Digital attacks directly perturb image pixels and mainly succeed in controlled experimental environments where full gradient access is available. In contrast, physical attacks must withstand real-world variations in lighting, viewpoint, camera resolution, printing distortion, and environmental noise. The adversarial pattern must remain effective after being printed and captured by cameras under diverse conditions, which significantly increases the difficulty of attack design. These challenges make physical adversarial attacks more complex yet more practical in real-world threat scenarios.

Many existing physical attacks optimize only a single submodule of the detector while neglecting the strong coupling among the classification head, regression head, and post-processing modules such as Non-Maximum Suppression (NMS). As a result, other modules compensate for the attack, reducing overall effectiveness. Some works use printable patches to minimize classification confidence~\cite{thys2019fooling}, while others print adversarial textures on clothing to target a single confidence output~\cite{xu2020adversarial}. However, these approaches often fail to simultaneously disrupt localization consistency and proposal suppression mechanisms. Furthermore, real-world attacks must handle complex variations in illumination, viewing distance, material reflectance, and pose deformation. Methods relying on transferable perturbations, such as universal patches~\cite{lee2019physical}, often remain unstable when deployed under dynamic physical conditions.


To address these challenges, we propose a novel physical adversarial attack method, TriPatch, which achieves efficient attacks against pedestrians through dynamic patch positioning and multi-objective joint optimization.  Our method features an adaptive patch generation strategy that uses semantic information from detection bounding boxes to localize patches over critical human body areas. Moreover, we employ a composite loss function, including target confidence suppression, bounding-box IoU penalty, and NMS suppression losses—to effectively attack feature extraction, localization, and post-processing modules. In addition, by integrating appearance consistency constraints, TriPatch preserves visual naturalness while substantially improving physical robustness.

\begin{figure*}[!t]
    \centering
    \includegraphics[scale=0.62]{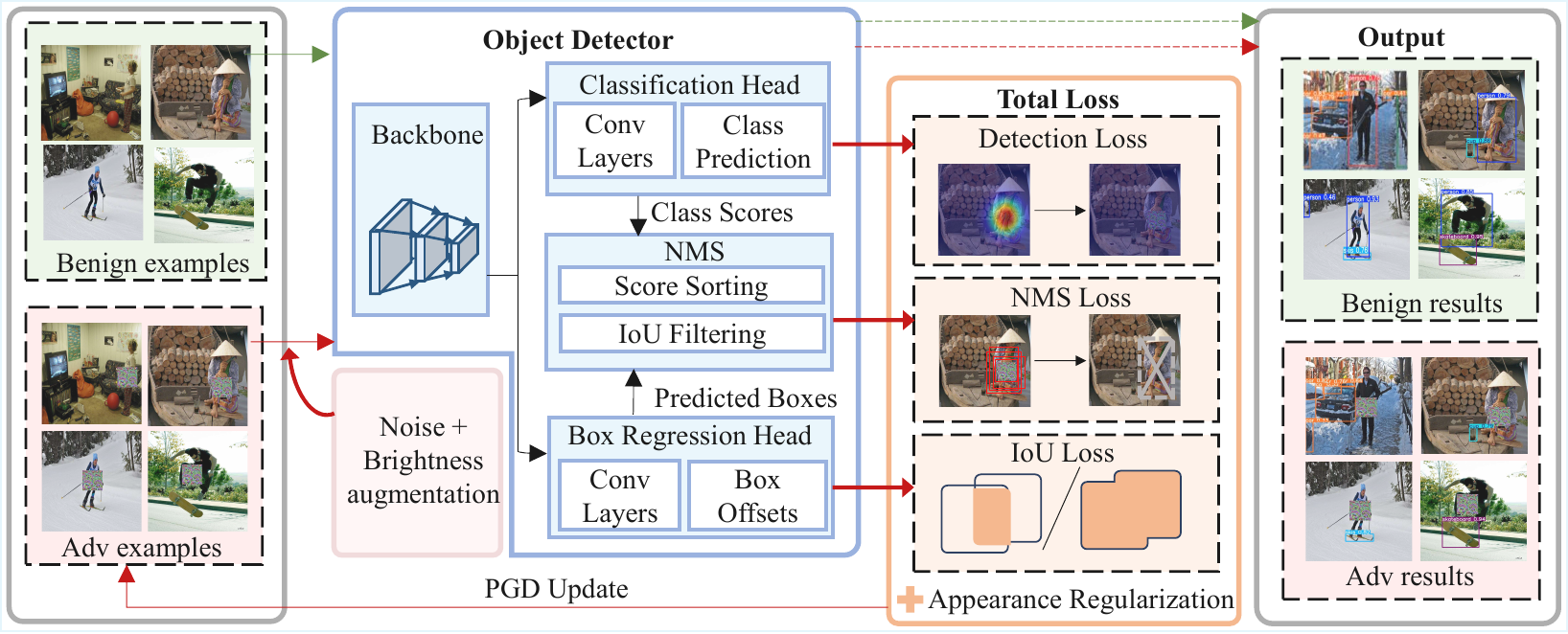}
    \caption{The pipeline of TriPatch.
    }
    \label{fig:pipeline}
\end{figure*}


Our main contributions are summarized as follows:
\begin{itemize}
    \item We propose an adversarial patch framework for physical-world pedestrian detection with robustness modules that enhance adaptability to environmental variations through color space.
    \item We design a triple-loss function that synchronizes attacks on key detector components and significantly improves success rates. 
    \item We conduct extensive experiments in digital and physical environments, showing that TriPatch greatly reduces mean Average Precision and effectively conceals pedestrians from detection.
\end{itemize}


%% file: Section/2-relatedwork.tex
\section{Related Works}
\subsection{Object Detectors}

Object detection aims to identify objects within an image and localize their corresponding bounding boxes. Existing approaches can be broadly categorized into two paradigms: two-stage detectors and one-stage detectors. Representative two-stage detectors, such as the R-CNN series (RCNN~\cite{girshick2014rcnn}, Fast RCNN~\cite{girshick2015fast}, and Faster RCNN~\cite{ren2015fasterrcnn}), achieve high detection accuracy by generating region proposals followed by classification and bounding box regression, though at the cost of computational efficiency. One-stage detectors such as the YOLO~\cite{redmon2016yolo,redmon2018yolov3,bochkovskiy2020yolov4,wang2023yolov7}, SSD~\cite{liu2016ssd}, and RetinaNet~\cite{lin2017focal} adopt end-to-end prediction and are well suited to real-time applications. The YOLO series continuously improves detection accuracy and inference speed through architectural enhancements and training strategies, evolving from the original YOLOv1 to more recent versions, and has been widely adopted in real-time scenarios such as autonomous driving and video surveillance. Modern object detectors have also benefited from advanced techniques including feature pyramid networks, attention mechanisms, and improved anchor design strategies. Despite these significant improvements in detection performance, the robustness and security aspects of these systems remain critical concerns, particularly when deployed in safety-critical applications.

\subsection{Digital Adversarial Attacks}

Digital adversarial attacks~\cite{zhou2024securely,advclip,zhou2023downstream,zhou2025numbod,zhou2024darksam,zhou2025sam2,yu2022towards, wang2024benchmarking, xia2024transferable,lu2026pretrain} aim to mislead model predictions by directly manipulating input pixel values, with early studies primarily concentrating on image classification tasks. ~\cite{szegedy2013intriguing} first revealed that imperceptible perturbations could cause deep neural networks to produce incorrect predictions and proposed an optimization-based L-BFGS method to craft adversarial examples. Building upon this finding, ~\cite{goodfellow2015explaining} introduced the Fast Gradient Sign Method (FGSM), which generates adversarial perturbations efficiently through a single-step gradient update, significantly reducing computational cost. To further enhance attack strength, ~\cite{madry2017towards} formulated the Projected Gradient Descent (PGD) attack as a multi-step iterative optimization procedure under norm constraints, establishing a strong first-order adversary benchmark in white-box settings. Beyond classification, digital attacks have been extended to more complex vision tasks such as object detection and semantic segmentation, where perturbations must simultaneously affect classification confidence and localization accuracy. As threat models diversified, black-box attacks also emerged, including transfer-based approaches that exploit surrogate models to improve cross-model generalization, as well as query-efficient and zero-shot generation methods that require limited or no access to the target model. Despite their effectiveness in controlled digital environments, these methods typically assume direct pixel-level manipulation and full gradient accessibility, leaving open challenges when confronting more constrained and structured attack scenarios.

\subsection{Physical Adversarial Patch Attacks}

Physical adversarial attacks can be categorized into texture-based and patch-based paradigms. Texture-based methods optimize global surface features of objects to create camouflage effects. For instance, researchers develop vehicle camouflage textures that blend with environmental backgrounds~\cite{zhang2019camou}, while others create 3D-printed turtle models with adversarial textures to fool classifiers~\cite{athalye2018synthesizing}. Recent advances in texture-based approaches focus on improving realism and physical robustness. Hu et al.~\cite{hu2023physically} propose physically realizable natural-looking clothing textures that evade person detectors via 3D modeling, demonstrating how adversarial textures can be seamlessly integrated into everyday clothing while maintaining attack effectiveness. Li et al.~\cite{Li_2025_ICLR} explore UV-attack methods for physical-world adversarial attacks against person detection via dynamic-NeRF-based UV mapping, showcasing the potential of neural rendering techniques for generating more sophisticated adversarial textures. Patch-based methods optimize perturbation patterns in specific regions on target surfaces. Early works propose printable patches placed near targets to deceive classifiers~\cite{brown2017adversarial}, while researchers design cardboard patches to conceal human detection via handheld placement~\cite{thys2019fooling}. For rigid objects, adversarial patterns are mapped onto 3D vehicle models to achieve multi-view attacks~\cite{wang2021dual}. For non-rigid targets, Thin-Plate Spline (TPS) transformations are employed to simulate fabric deformations, generating wrinkle-resistant adversarial T-shirts~\cite{xu2020adversarial}. Recent patch-based research emphasizes improving stealthiness and robustness: researchers leverage Generative Adversarial Networks (GANs) to synthesize natural-texture patches~\cite{hu2021naturalistic} and introduce visual consistency loss to constrain color distributions~\cite{duan2020adversarial}. Wei et al.~\cite{Wei_2024_NeurIPS} revisit adversarial patches for designed camera-agnostic attacks against person detection, Guesmi et al.~\cite{guesmi2024dap} develop dynamic adversarial patches for evading person detectors, and Tan et al.~\cite{Tan_2024_TIFS} propose dynamically optimized ensemble models for adversarial patch generation.
Despite these advances, existing methods often struggle with limited attack success rates in challenging real-world conditions and insufficient robustness against physical variations such as lighting changes, viewing angles, and environmental disturbances, highlighting the need for more comprehensive approaches that can achieve consistently high effectiveness across diverse practical scenarios.

%% file: Section/3-methodology.tex
\section{Methodology}\label{sec:methodology}

\subsection{Problem Formulation}\label{sec:problem}

Object detection is a fundamental task in computer vision that aims to localize and classify objects within images by outputting candidate bounding boxes, category labels, and confidence scores. Given an input image $x \in \mathbb{R}^{H \times W \times 3}$, where $H$ and $W$ represent the height and width of the image respectively, a detector $f(\cdot)$ produces a set of predictions $f(x) = \{(B_n, Y_n, c_n)\}_{n=1}^N$, where $B_n$ denotes the coordinates of the $n$-th bounding box, $Y_n \in \mathcal{Y}$ is the corresponding category label from the predefined class set $\mathcal{Y}$, and $c_n \in [0,1]$ is the classification confidence score indicating the model's certainty about the prediction. The parameter $N$ denotes the total number of candidate predictions returned by the detector on the image $x$, which typically varies depending on the complexity and content of the input image.

For pedestrian detection specifically, we focus on the "person" class within the object detection framework. The ground-truth bounding boxes for pedestrian instances are denoted as $\mathcal{B}^*(x) = \{B_m^*\}_{m=1}^{M(x)}$, where $B_m^*$ represents the $m$-th ground-truth pedestrian bounding box with its center coordinates and dimensions, and $M(x)$ denotes the total number of ground-truth pedestrians present in image $x$.

\noindent\textbf{Threat model.}
Our goal is to design a printable adversarial patch $p \in \mathbb{R}^{H_p \times W_p \times 3}$ that can be physically attached to a human subject to evade pedestrian detection systems. The patch dimensions are constrained by $H_p$ and $W_p$, which represent the height and width of the patch respectively, with the realistic constraints $H_p < H$ and $W_p < W$ to ensure the patch remains inconspicuous and practical for real-world deployment. The adversarial sample is generated by overlaying the patch $p$ onto the input image $x$ at a specific location, expressed as $x^{adv} = x \oplus p$, where $\oplus$ denotes the patch application operator that performs spatial transformation and blending of the patch onto the original image. This operator encompasses the geometric transformations necessary to position the patch appropriately on the human body, including scaling, rotation, and perspective adjustments to account for varying viewpoints and distances.

The adversary operates under a black-box threat model, where the attacker has no knowledge of the internal architecture, parameters, or training data of the target detection system. The adversary can only observe the final detection outputs, making this a realistic and challenging attack scenario. The primary objective is to suppress the confidence scores of pedestrian detections such that the detector fails to output valid predictions for the person class, effectively rendering the pedestrian invisible to the detection system. This stealth requirement is critical for applications where the adversary seeks to avoid detection while maintaining normal appearance.

The optimization objective for generating effective adversarial patches is formulated as:
\begin{equation}
\min_{p}\; \mathbb{E}_{x \sim \mathcal{D}}
\left[ \operatorname{mean}_{\,n:\, Y_n^{adv} = \text{person}} \; c_n^{adv} \right],
\label{eq:TriPatch_define}
\end{equation}
where $\mathcal{D}$ represents the distribution of input images containing pedestrians, and $f(x^{adv}) = \{(B_n^{adv}, Y_n^{adv}, c_n^{adv})\}_{n=1}^N$ denotes the detector's output on the adversarial image $x^{adv}$. The expectation is taken over all possible images in the dataset to ensure the patch generalizes across different scenarios, lighting conditions, and pedestrian poses. By minimizing the mean confidence score $c_n^{adv}$ of all predicted pedestrian instances (where $Y_n^{adv} = \text{person}$), the adversarial patch reduces the likelihood that the detector recognizes and correctly classifies the pedestrian, ultimately leading to detection failure when confidence scores fall below the detection threshold $\tau_{det}$, typically set around 0.5 in most detection systems.

\subsection{Triple-Loss Mechanism}
Motivated by the observation that modern pedestrian detectors rely on multiple coupled cues rather than a single score, we design our objective from a decomposition perspective. Specifically, an effective physical patch should simultaneously affect the detector's confidence estimation, localization consistency, and the structured proposal interactions induced by post-processing. Therefore, instead of optimizing a single attack term, we formulate a triple-loss mechanism that jointly targets these complementary aspects, leading to more stable adversarial behavior under real-world imaging variations.

The rationale behind our multi-objective approach stems from the inherent redundancy in modern detection architectures. When attacking only the classification confidence, the regression head can still provide accurate localization information, allowing post-processing modules to recover valid detections. Similarly, targeting only bounding box regression may leave classification scores intact, enabling the detector to maintain high confidence predictions. By simultaneously disrupting all three components—classification, regression, and post-processing—our method ensures comprehensive attack coverage that prevents compensatory mechanisms from maintaining detection performance.

To achieve pedestrian invisibility in the physical world, we propose a triple-loss mechanism that synergistically attacks vulnerabilities in detection models. The pipeline of TriPatch is depicted in \Cref{fig:pipeline}. We optimize on the adversarial image $x^{adv}$ a total objective composed of three primary losses and one appearance regularizer:
\begin{equation}
L_{\text{total}}
=  \lambda_{\text{det}} L_{\text{det}}
+  \lambda_{\text{iou}} L_{\text{iou}}
+  \lambda_{\text{nms}} L_{\text{nms}}
+  \lambda_{\text{app}} L_{\text{app}}, 
\label{eq:L_total}
\end{equation}
where $\lambda_{\text{det}}$, $\lambda_{\text{iou}}$, $\lambda_{\text{nms}}$, and $\lambda_{\text{app}}$ are weighting coefficients that balance the contribution of each loss component. These hyperparameters allow fine-tuning the attack strategy based on the specific detector architecture and deployment requirements.

\noindent\textbf{Detection confidence loss.} To suppress the visibility of the pedestrian class, we minimize the confidence scores among all pedestrian candidates. This loss directly targets the classification head of the detector:
\begin{equation}
L_{\text{det}}
= \mathbb{E}_{x\sim\mathcal{D}}
\!\left[
\operatorname{mean}_{\,n \in \mathcal{S}(x^{adv})}\; c_n^{adv}
\right],
\label{eq:L_det}
\end{equation}
where $\mathcal{S}(x^{adv})=\{\,n \mid Y_n^{adv}=\text{person}, \, c_n^{adv} > \tau_{\text{conf}}\,\}$ is the index set of candidates predicted as person on $x^{adv}$ with confidence above a threshold $\tau_{\text{conf}}$. This formulation focuses the attack on high-confidence predictions that are most likely to survive post-processing, making the optimization more targeted and efficient. The expectation over the dataset distribution $\mathcal{D}$ ensures that the patch generalizes across diverse pedestrian appearances and environmental conditions.

\noindent\textbf{Bounding box loss.} To reduce spatial alignment between high-confidence candidates and the ground-truth pedestrian boxes, we penalize the Intersection over Union (IoU) weighted by confidence for candidates that overlap with ground truth annotations: 
\begin{equation}
L_{\text{iou}}
= \frac{1}{|\mathcal{S}(x^{adv})||\mathcal{B}^*(x)|}
\sum_{n\in\mathcal{S}(x^{adv})}
\sum_{m=1}^{M(x)}
\mathrm{IoU}(B_n^{adv},B_m^*)\cdot c_n^{adv},
\label{eq:L_iou}
\end{equation}
where $\mathrm{IoU}(B_n^{adv},B_m^*)$ computes the intersection over union between the predicted bounding box $B_n^{adv}$ and the ground-truth box $B_m^*$. This loss encourages the regression head to produce inaccurate localizations, either by shifting predicted boxes away from the true pedestrian location or by altering their size and aspect ratio. The confidence weighting $c_n^{adv}$ prioritizes the disruption of high-confidence predictions, as these are more likely to be retained as final detections.

\begin{figure*}[t]
    \centering
    \includegraphics[scale=0.68]{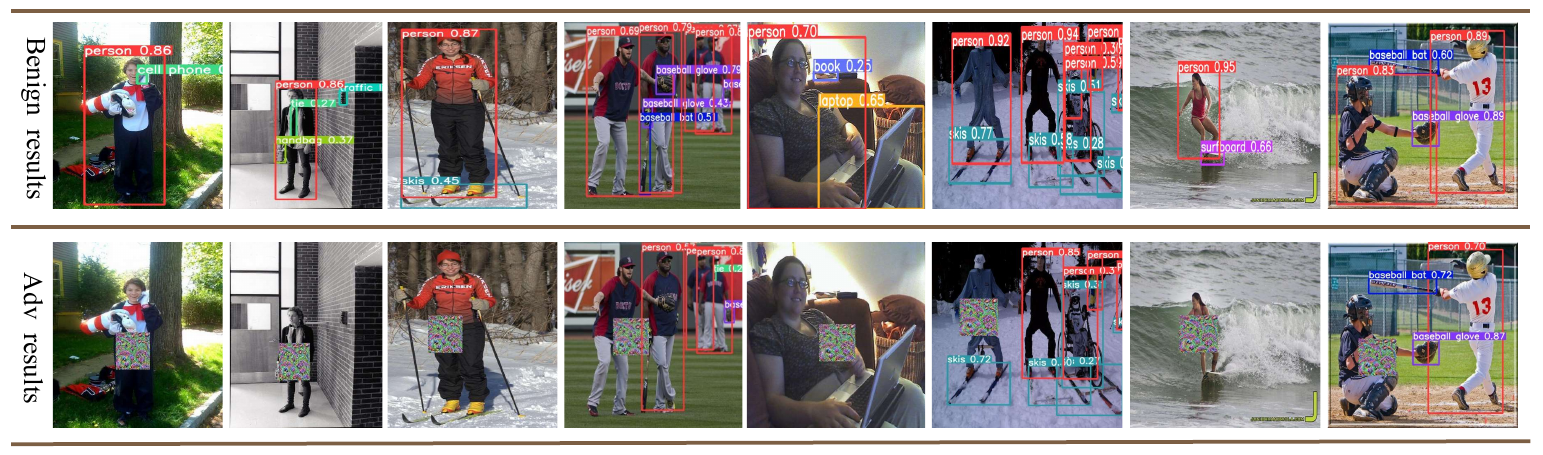}
    \caption{Digital-World Adversarial Attack Results of Our Method.
    }
    \label{Fig:Digital}
       \vspace{-0.4cm}
\end{figure*}

\noindent\textbf{NMS loss.} To disrupt the spatial clustering of high-score candidates and facilitate their removal by non-maximum suppression, we penalize pairwise overlaps among the top-$K$ confident boxes: 
\begin{equation}
L_{\text{nms}}
= \frac{1}{|\mathcal{P}_K|}\!
\sum_{(i,j)\in \mathcal{P}_K}
\phi\!\left(\mathrm{IoU}\!\left(B_i^{adv}, B_j^{adv}\right) - \tau_{\mathrm{nms}}\right)\,
c_i^{adv}\, c_j^{adv},
\label{eq:L_nms}
\end{equation}
where $\mathcal{P}_K = \{(i,j) \mid i \neq j, \, i,j \in \text{top-}K\}$ is the set of ordered pairs from the top-$K$ confidence-ranked boxes, $\tau_{\mathrm{nms}}$ is the NMS overlap threshold (typically 0.5), and $\phi(u)=\log(1+e^{u})$ is a smooth penalty function that ensures differentiability for gradient-based optimization. This term encourages high-score boxes to be spatially dispersed or to have overlaps that exceed the NMS threshold, increasing the likelihood that they will suppress each other during post-processing.

\noindent\textbf{Appearance regularization.}
To improve printability and robustness across different environmental conditions, we constrain the statistical properties of the patch to ensure it remains within realistic color ranges:
\begin{equation}
L_{\text{app}}
= \big(\mu_p - 0.5\big)^2 \;+\; \max\!\left(0,\, \sigma_{\min} - \sigma_p \right) + \lambda_{\text{smooth}} \|\nabla p\|_2^2,
\label{eq:L_app}
\end{equation}
where $\mu_p = \frac{1}{H_p \times W_p \times 3}\sum_{i,j,k} p_{i,j,k}$ and $\sigma_p^2 = \frac{1}{H_p \times W_p \times 3}\sum_{i,j,k} (p_{i,j,k} - \mu_p)^2$ are the global mean and variance of patch $p$ computed over all spatial locations and RGB channels, $\sigma_{\min}$ is the minimum variance threshold to prevent over-smoothing, $\nabla p$ represents the spatial gradient of the patch, and $\lambda_{\text{smooth}}$ controls the smoothness regularization strength. The first term prevents extreme brightness values that may be difficult to print or appear unnatural, the second term maintains sufficient color variation to preserve attack effectiveness, and the third term promotes spatial smoothness to reduce conspicuousness.

\noindent\textbf{Data augmentation.}
To address environmental disturbances and device imaging variations in physical deployment, we adopt comprehensive data augmentation strategies during training that simulate real-world conditions:
\begin{equation}
\begin{split}
x_{\text{aug}}^{adv} &= \mathrm{clip}(\alpha \cdot \mathrm{transform}(x^{adv}) + \eta,\,0,1),
\end{split}
\label{eq:data_aug}
\end{equation}
where $\alpha \sim \mathcal{U}[1-\delta_b,\, 1+\delta_b]$ controls brightness scaling with $\delta_b$ setting the brightness jitter range, $\eta \sim \mathcal{N}(0,\, \sigma_{\text{aug}}^{2}\mathbf{I})$ represents additive Gaussian noise with standard deviation $\sigma_{\text{aug}}$, $\mathrm{transform}(\cdot)$ applies geometric transformations including rotation, scaling, and perspective distortion, and $\mathrm{clip}(\cdot,0,1)$ clamps pixel values to the valid range $[0,1]$. These augmentations simulate the natural variations encountered in real-world deployment, including changes in lighting conditions, camera angles, distances, and environmental factors such as shadows and reflections.


%% file: Section/4-experiments.tex
\begin{table*}[!t]
\setlength{\abovecaptionskip}{4pt}
  \centering
      \caption{Attack performance in terms of mAP on INRIA and MS-COCO datasets using different detectors.}
   \scalebox{1.0}{
    \begin{tabular}{cc*{8}{c}}  
\toprule[1.5pt]
\multirow{2}{*}{Dataset} & \multirow{2}{*}{Victim} & \multicolumn{8}{c}{Trained on} \\
\cmidrule(lr){3-10}
 &  & YOLOv2 & YOLOv3 & YOLOv3tiny & YOLOv4 & YOLOv5 & YOLOv8 & YOLOv9 & Faster RCNN \\
\midrule
    \multirow{8}{*}{\rotatebox{90}{INRIA}} & YOLOv2 & 0.89 & 14.71 & 7.27 & 37.16 & 4.99 & 12.35 & 15.67 & 44.25 \\
    & YOLOv3  & 0.45 & 11.98 & 5.73 & 21.64 & 10.86 & 18.42 & 22.15 & 37.60 \\
    & YOLOv3tiny  & 0.25 & 9.42 & 2.15 & 29.32 & 5.73 & 14.28 & 17.93 & 32.52 \\
    & YOLOv4  & 0.08 & 10.58 & 4.76 & 18.87 & 5.70 & 16.74 & 20.31 & 34.30 \\
    & YOLOv5  & 0.29 & 13.44 & 9.72 & 31.92 & 6.19 & 19.85 & 23.47 & 32.21 \\
    & YOLOv8  & 0.34 & 14.83 & 9.45 & 33.27 & 8.94 & 7.42 & 25.68 & 34.92 \\
    & YOLOv9  & 0.31 & 13.29 & 8.18 & 31.64 & 7.76 & 12.41 & 6.95 & 32.75 \\
    & Faster RCNN  & 1.15 & 11.49 & 5.60 & 36.86 & 10.21 & 20.67 & 14.83 & 28.40 \\
\midrule
    \multirow{8}{*}{\rotatebox{90}{COCO}} & YOLOv2 & 0.59 & 3.99 & 3.69 & 8.47 & 1.19 & 4.82 & 6.15 & 13.96 \\
    & YOLOv3  & 8.72 & 1.33 & 4.05 & 26.29 & 7.95 & 12.34 & 14.67 & 21.62 \\
    & YOLOv3tiny  & 2.62 & 4.38 & 0.40 & 11.96 & 2.12 & 6.28 & 7.91 & 16.94 \\
    & YOLOv4  & 7.29 & 5.42 & 3.17 & 11.25 & 3.16 & 8.74 & 10.31 & 18.15 \\
    & YOLOv5  & 10.62 & 8.26 & 5.67 & 25.02 & 3.55 & 11.85 & 13.47 & 21.75 \\
    & YOLOv8  & 6.43 & 9.18 & 7.25 & 19.34 & 5.87 & 2.78 & 8.23 & 24.28 \\
    & YOLOv9  & 5.76 & 8.51 & 6.48 & 17.72 & 4.91 & 6.45 & 2.52 & 22.64 \\
    & Faster RCNN  & 11.23 & 5.34 & 4.47 & 29.82 & 9.70 & 14.87 & 16.25 & 8.31 \\
    \bottomrule[1.5pt]
    \end{tabular}%
    }
  \label{tab:performance}%
\end{table*}%

\begin{figure*}[!t]
    \centering
    \includegraphics[scale=0.62]{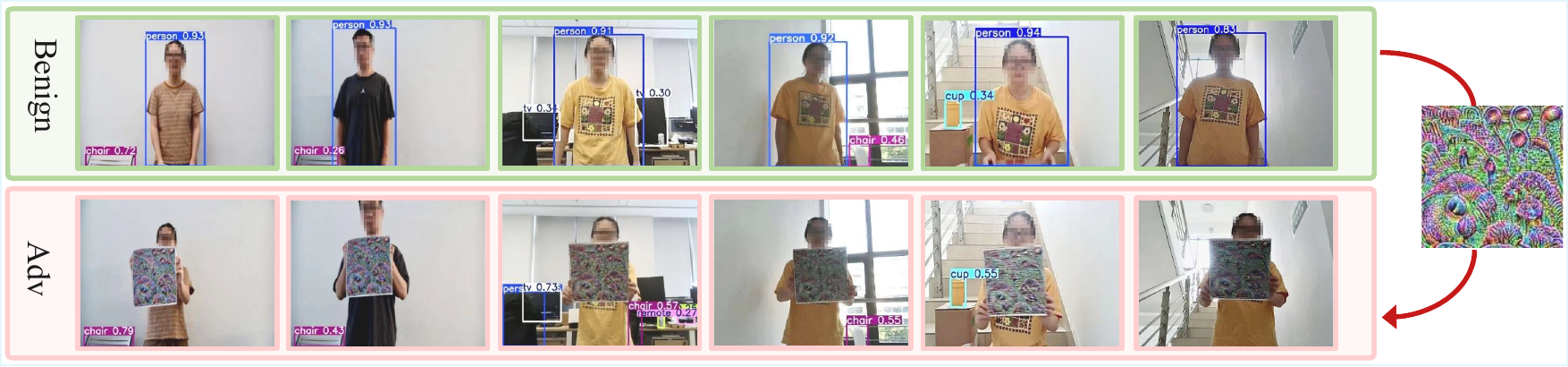}
    \caption{Physical attack demo of TriPatch.}
    \label{Fig:physical}
\end{figure*}
\section{Experiments}
\subsection{Experimental Setup}
\label{sec:experimental_setup}

\noindent\textbf{Datasets and models.} 
We evaluate the attack effectiveness of our proposed method using two benchmark datasets: INRIA~\cite{dalal2005hog} and MS-COCO~\cite{lin2014coco}. For the object detection models, we select a diverse range of detectors, including YOLOv2~\cite{redmon2016yolo9000}, YOLOv3~\cite{redmon2018yolov3},YOLOv3tiny, YOLOv4~\cite{bochkovskiy2020yolov4}, YOLOv5~\cite{jocher2020yolov5}, YOLOv8~\cite{YOLOv8_2023_GitHub}, YOLOv9~\cite{Wang_2024_ECCV} and Faster RCNN~\cite{ren2015fasterrcnn}.

\noindent\textbf{Parameter setting.}
In our experiments, the learning rate is set to 0.02, the batch size is configured to 8, and the training is conducted for a total of 25 epochs. The input image resolution is set to 640 × 640, while the output image resolution is standardized to 128 × 128. These hyperparameters are chosen following common practice in prior adversarial patch studies to ensure fair comparison and stable optimization.

\noindent\textbf{Evaluation metrics.}
We adopt mean Average Precision (mAP) as the primary evaluation metric. 
Following the protocol established in prior work~\cite{thys2019fooling,hu2021naturalistic}, we treat the detection outputs on clean samples without any adversarial perturbations as the ground truth boxes, for which the mAP is assumed to be 100\%. Under this definition of ground truth, we compute the Average Precision (AP) in the presence of adversarial patches.

\subsection{Digital Attack Performance} \label{sec:digital_attack_performance}


To comprehensively evaluate the attack performance of the proposed method, we conduct experiments on two benchmark datasets, INRIA and MS-COCO, utilizing seven mainstream object detectors. The evaluation metric employed is mean Average Precision (mAP), which serves as the primary indicator of detection accuracy by measuring the area under the precision-recall curve across different IoU thresholds. 
As reported in \Cref{tab:performance}, the proposed method consistently leads to significantly lower mAP scores across various detector configurations, demonstrating its strong effectiveness in compromising object detection performance. The experimental results reveal remarkable attack success, with our method achieving extremely low mAP scores when targeting the same detector architecture used for training. For instance, patches trained on YOLOv9 reduce the victim detector's mAP to merely 6.95\% on INRIA dataset, while YOLOv3tiny-trained patches achieve 0.40\% on MS-COCO dataset. Even when attacking different architectures, our method maintains substantial effectiveness, consistently reducing mAP scores below 45\% across all configurations. 
\Cref{Fig:Digital} presents several examples of the adversarial patches generated by our method. 
We apply the YOLOv5 detector to both clean images without patches  and images modified with patches produced by our method. 
The results demonstrate that when the patches generated by our method are applied to human subjects in dataset images, the detection bounding boxes for the patch-wearing individuals disappear, rendering them undetectable by object detection systems, thereby confirming the success of the attack.

\subsection{Transferability Study} \label{sec:transferability}


To comprehensively evaluate the transferability of our proposed TriPatch method, we conduct extensive cross-detector experiments using both INRIA and MS-COCO datasets. The results are summarized in \Cref{tab:performance}, where the victim detectors used during testing are shown on the horizontal axis, and the detectors used for training the adversarial patches are shown on the vertical axis. This experimental design allows us to assess how well adversarial patches trained on one detector can transfer to attack other detectors, which is crucial for understanding the practical applicability of the method in real-world scenarios where the target detector architecture may be unknown.

The experimental results demonstrate that TriPatch consistently achieves significant performance degradation across various detector combinations, indicating strong transferability and robustness. The method shows excellent transferability across different YOLO architectures, suggesting that the learned adversarial patterns effectively exploit common architectural features shared among YOLO variants. More importantly, the method demonstrates strong cross-architecture transferability, with patches trained on YOLO-based detectors successfully transferring to Faster RCNN, and vice versa.

\begin{table*}[!t]
\setlength{\abovecaptionskip}{4pt}
  \centering
      \caption{The mAP of comparison study.}
    \scalebox{1.0}{
    \setlength{\tabcolsep}{10pt} 
    \begin{tabular}{c*{8}{c}}
    \toprule[1.5pt]
    \multirow{2}[4]{*}{Method} & \multicolumn{8}{c}{Detector} \\
\cmidrule(lr){2-9}
& YOLOv2 & YOLOv3 & YOLOv3tiny & YOLOv4 & YOLOv5 & YOLOv8 & YOLOv9 & Faster RCNN \\
    \midrule
    AdvPatch~\cite{thys2019fooling} & 5.66 & 13.89 & 8.74 & 19.67 & 13.39 & 16.24 & 15.78 & 39.41 \\
    T-SEA~\cite{huang2023tsea} & 1.73 & 14.17 & 4.28 & 23.08 & 8.76 & 12.35 & 11.89 & 31.16 \\
    UPC~\cite{huang2019upc} & 48.62 & 54.4 & 63.82 & 64.21 & 53.67 & 58.91 & 57.63 & 61.87 \\
    NAP~\cite{hu2021naturalistic} & 12.06 & 34.93 & 10.02 & 22.63 & 19.83 & 25.47 & 24.12 & 42.47 \\
    CAP~\cite{Wei_2024_NeurIPS} & 3.25 & 13.52 & 6.18 & 21.15 & 10.84 & 14.67 & 13.92 & 35.73 \\
    Ours & 0.89 & 11.98 & 2.15 & 18.87 & 6.19 & 7.42 & 6.95 & 28.40 \\
    
    \bottomrule[1.5pt]
    \end{tabular}%
    }
  \label{tab:compare}%
       \vspace{-0.2cm}
\end{table*}%

\begin{figure*}[!t]  
\setlength{\abovecaptionskip}{4pt}
  \centering

  \subcaptionbox{Epoch}{\includegraphics[height=3.1cm]{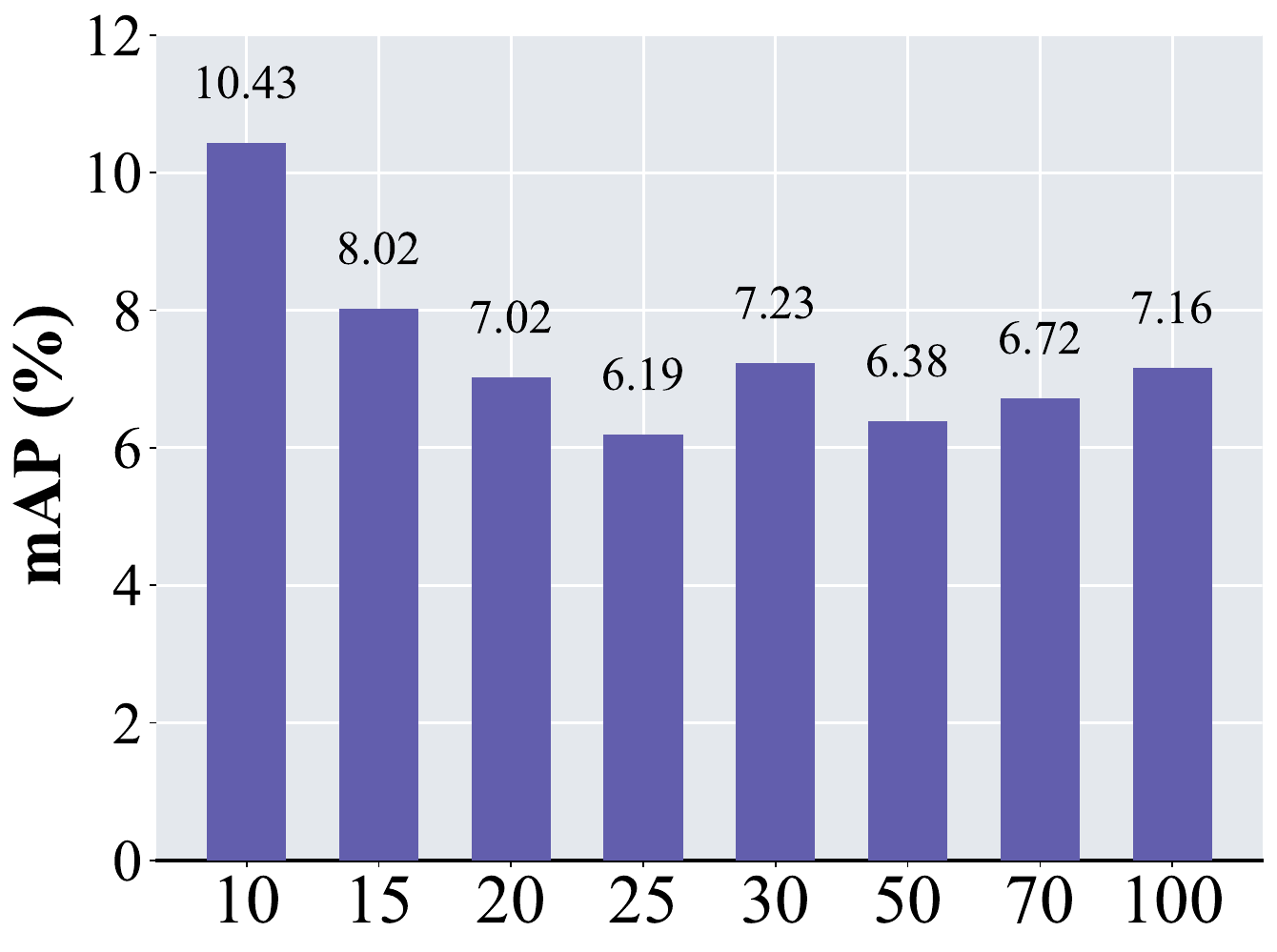}}
  \hfill
  \subcaptionbox{Patch size}{\includegraphics[height=3.1cm]{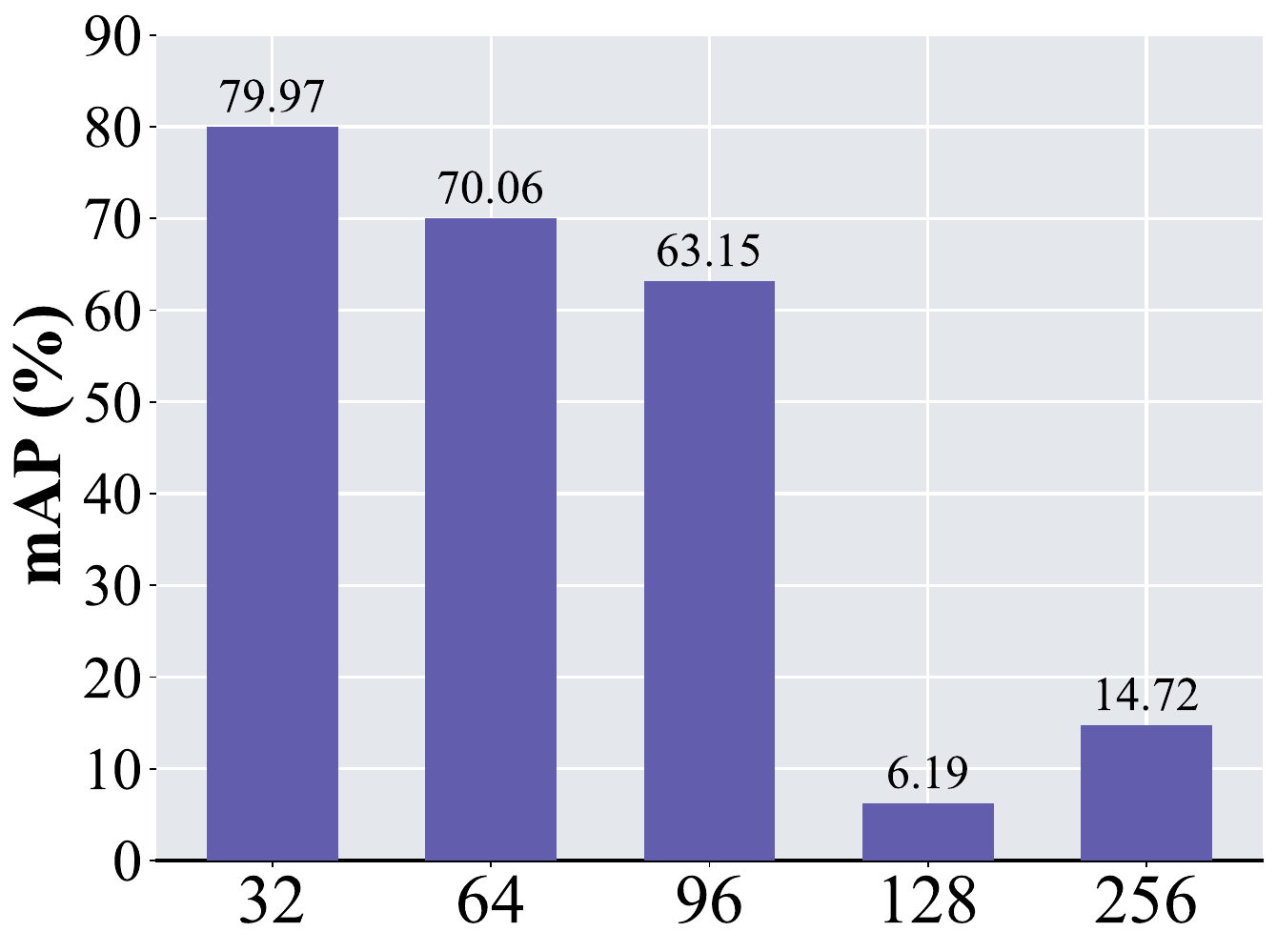}}
  \hfill
  \subcaptionbox{Module}{\includegraphics[height=3.1cm]{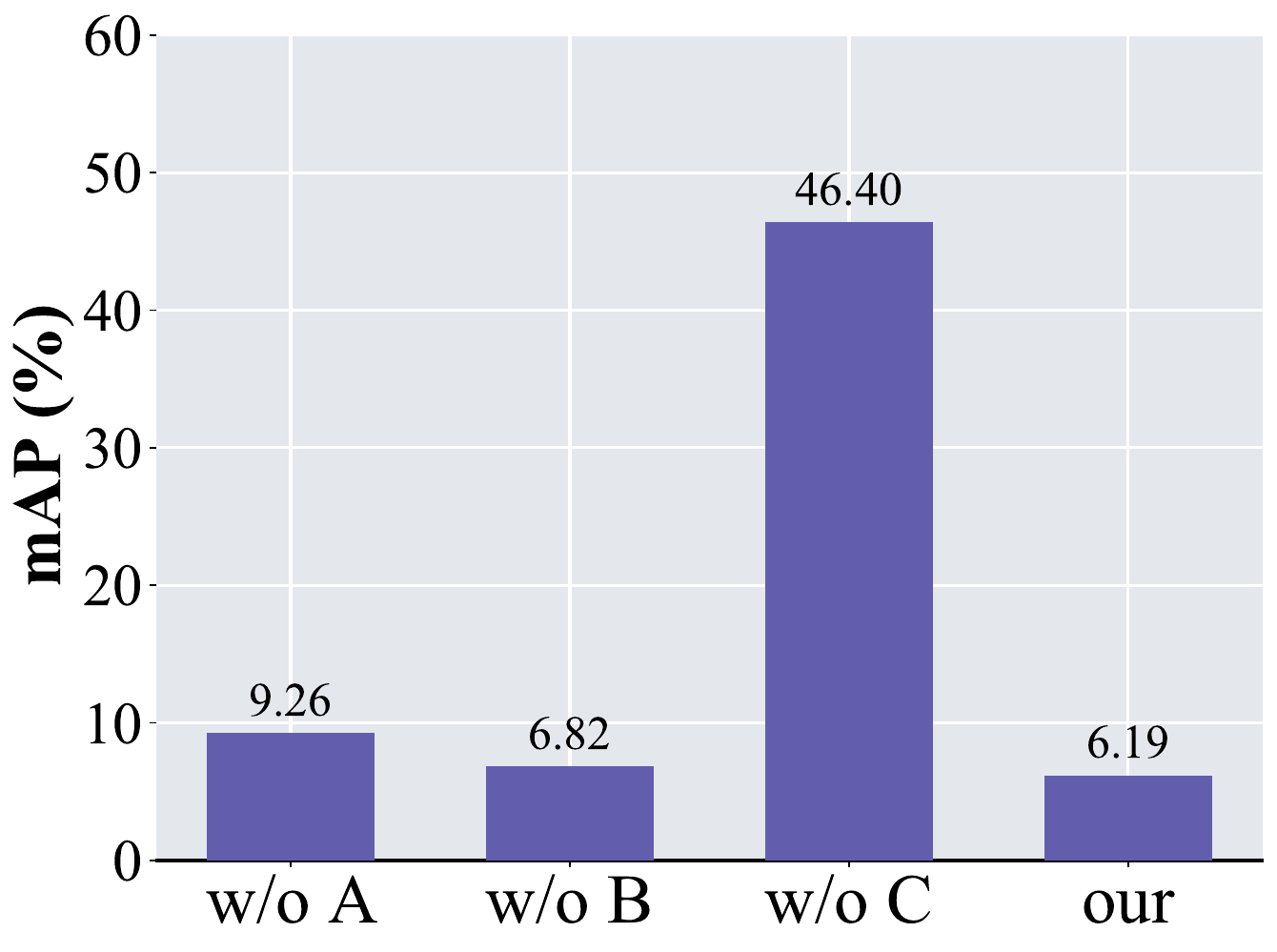}}
  \hfill
  \subcaptionbox{Loss weights}{\includegraphics[height=3.1cm]{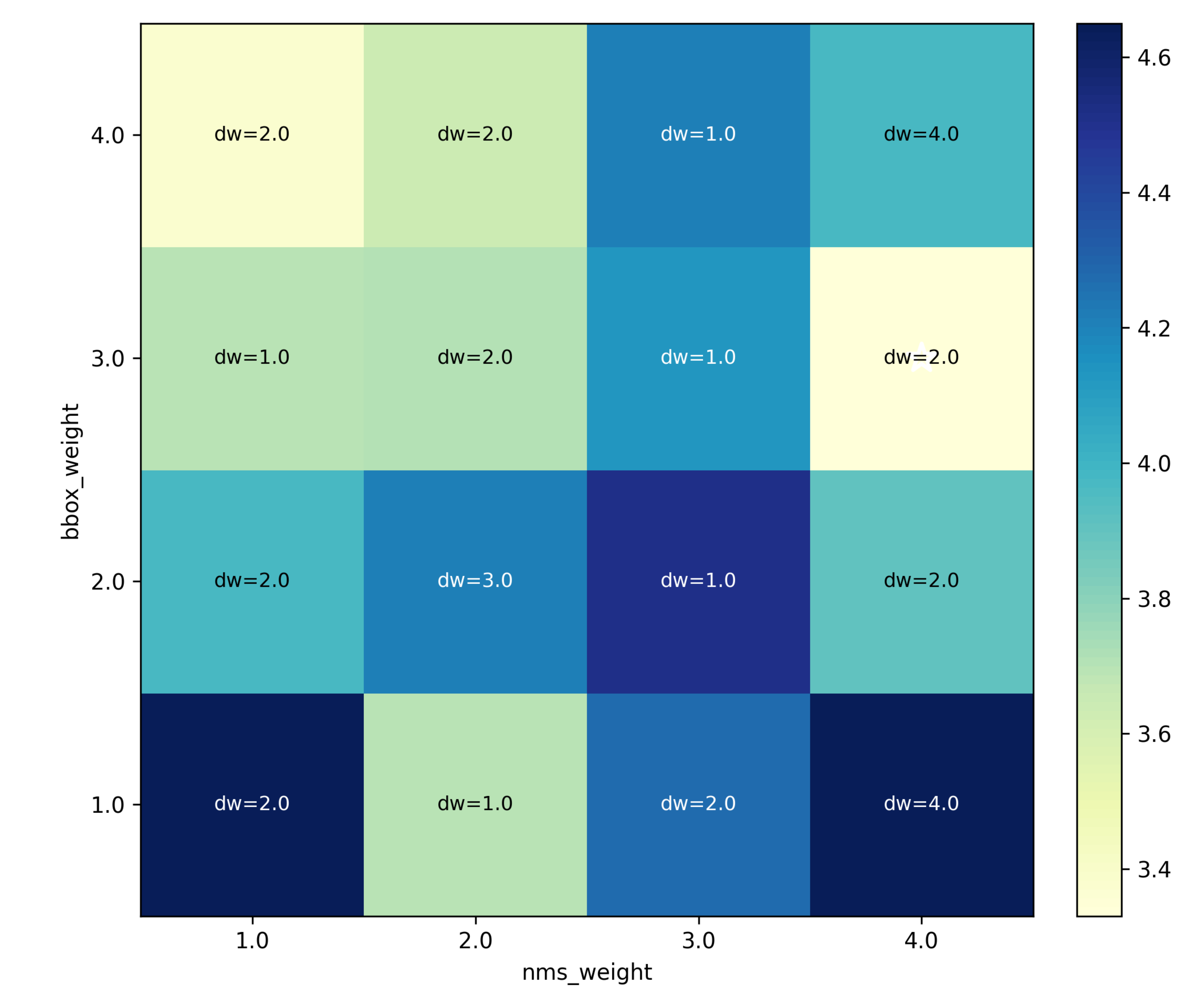}}

  \caption{The results (\%) of ablation study. (a) - (d) investigate the effect of different epochs, patch sizes, modules and loss weights respectively.}
  \label{Fig:ablation_results}
\end{figure*}

The results reveal interesting dataset-dependent patterns, with attacks generally being more effective on the INRIA dataset compared to MS-COCO. This can be attributed to the different characteristics of these datasets, where INRIA primarily focuses on pedestrian detection with relatively simple backgrounds, while MS-COCO contains more diverse object categories and complex scenes. Additionally, the method shows consistent effectiveness across different versions of the same detector series, demonstrating that fundamental vulnerabilities persist across architectural improvements.

\subsection{Comparison Study} \label{sec:compare}

To further validate the effectiveness of our approach, we compare it against five widely adopted adversarial attack methods: AdvPatch~\cite{thys2019fooling}, T-SEA~\cite{huang2023tsea}, NAP~\cite{hu2021naturalistic}, UPC~\cite{huang2019upc} and CAP~\cite{Wei_2024_NeurIPS}. These baseline methods represent diverse attack strategies and design philosophies in the field of adversarial attacks against object detection systems.  We select INRIA as the attacked dataset, with model and backbone settings as described in \Cref{sec:experimental_setup}. The comparative mAP results are presented in \Cref{tab:compare}.

The experimental results demonstrate that our proposed method achieves superior attack performance across all tested detectors. Our TriPatch approach consistently outperforms existing methods, achieving the lowest mAP scores in most cases. Particularly noteworthy is the performance on YOLOv2, where our method reduces the mAP to just 0.89\%, demonstrating exceptional attack effectiveness. On the YOLO series of detectors, our method maintains consistently strong performance with mAP scores ranging from 0.89\% to 18.87\%. Even on the more robust Faster RCNN detector, our approach achieves 28.40\% mAP, significantly outperforming most baseline methods. These results validate the effectiveness of our multi-patch collaborative attack strategy and demonstrate the superiority of the TriPatch approach in compromising various object detection architectures.

\subsection{Physical Attack Performance} \label{sec:physical_attack_performance}

To comprehensively evaluate the effectiveness of our proposed method in real-world physical environments, we conduct extensive physical attack experiments. Visual experimental results are demonstrated in \Cref{Fig:physical}, showcasing the attack performance of printed patches across diverse real-world scenarios.

We conduct adversarial evaluations on human targets across various environmental conditions, including indoor office environments, corridor passages, and locations with complex backgrounds. In the experiments, volunteers wear adversarial patches generated by our method while performing normal activities within the surveillance range of object detection systems. Comparative experiments reveal that under benign conditions, where no adversarial patch is worn, the object detection system accurately identifies and localizes human targets, with detection bounding boxes clearly marking the human regions. However, when volunteers wear our generated adversarial patches, the object detection system completely fails, unable to detect the presence of humans, effectively achieving an "invisibility" effect. These results conclusively demonstrate that patches generated by our method can successfully attack object detectors, effectively rendering persons invisible to detection systems in real-world physical environments.

\begin{figure}[!t]
    \centering
    \includegraphics[scale=0.28]{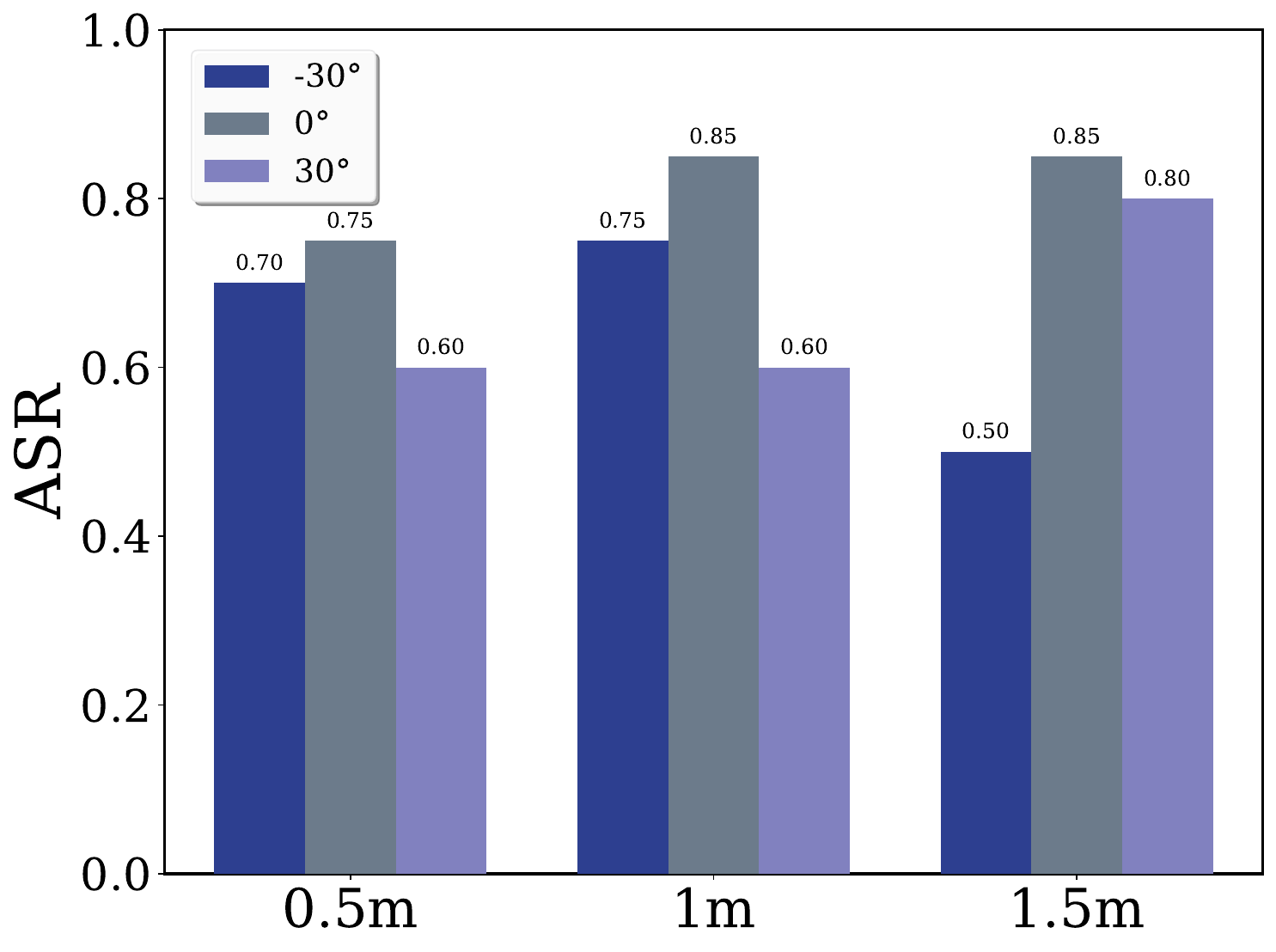}
    \caption{Physical attack performance at different angles and distances.}
    \label{Fig:physical_angles}
\end{figure}

To further quantify the robustness of physical attacks, we conduct systematic quantitative experiments using printed adversarial patches with dimensions of 24 cm × 24 cm. As shown in \Cref{Fig:physical_angles}, we evaluate two critical factors that influence the effectiveness of physical adversarial attacks under controlled variable conditions: viewing distance and viewing angle. Specifically, the viewing distance refers to the straight-line distance between the human target wearing the adversarial patch and the camera. We test three different distances: 0.5 m, 1 m, and 1.5 m, which cover typical near-range, medium-range, and far-range observation conditions in surveillance scenarios. The viewing angle represents the angle between the human body's frontal normal vector and the camera's optical axis. We select three representative angles: -30°, 0°, and +30°, where 0° indicates the human body facing directly toward the camera, and ±30° simulate common lateral pose variations encountered in surveillance areas.

For each distance-angle configuration combination, we conduct 20 independent experiments and calculate the Attack Success Rate (ASR) as the evaluation metric. The experimental results demonstrate that the proposed adversarial patch maintains high attack success rates across different observation conditions. These quantitative results reveal several important performance characteristics. First, the adversarial patch exhibits stronger attack effectiveness at moderate distances (1-1.5 m), indicating that the patch possesses robustness to distance variations and does not significantly degrade in attack performance due to moderate changes in distance. Second, across different viewing angles, the patch maintains considerable attack success rates, particularly achieving consistently high performance at the 0° frontal angle, while also reaching acceptable attack effectiveness at ±30° lateral angles. This demonstrates good tolerance to common pose and camera viewpoint deviations encountered in real-world environments. Overall, these findings provide quantitative evidence for the effectiveness and stability of the proposed method in the physical world, proving that the method is not limited to digital simulation scenarios but can achieve effective and relatively stable adversarial attacks in real physical environments.

\subsection{Ablation Study}

In this section, we examine the impact of different modules on the overall performance of the proposed method. The evaluation is conducted on the INRIA dataset using the YOLOv5 detector, focusing on the contribution of each component to the effectiveness of the TriPatch attack framework.

\noindent\textbf{Effect of iteration count.} \Cref{Fig:ablation_results} (a) illustrates the relationship between the number of optimization iterations and attack strength. Overall, the results demonstrate that more iterations consistently lead to improved attack effectiveness.

\noindent\textbf{Effect of patch size.} 
To investigate the influence of patch size relative to pedestrian scale on attack performance, we conducted systematic experiments on the INRIA dataset. As shown in \Cref{Fig:ablation_results} (b), larger patch sizes yield stronger attack performance, which aligns with our theoretical expectations.

\noindent\textbf{Effect of the loss function.} As illustrated in \Cref{Fig:ablation_results} (c), we perform an ablation study on each individual loss component. We use A, B and C to represent $\mathcal{L}_{\text{det}}$, $\mathcal{L}_{\text{iou}}$ and $\mathcal{L}_{\text{nms}}$ respectively. The results indicate that removing any single term leads to performance degradation, suggesting that all loss components are essential and work synergistically within the TriPatch framework.

\noindent\textbf{Effect of loss weights.} As shown in \Cref{Fig:ablation_results} (d), we ablate the weights of the three loss terms: detection confidence suppression loss (detect\_weight), bounding box disruption loss (bbox\_weight), and NMS circumvention loss (nms\_weight), and visualize the resulting Attack mAP. Lower values indicate stronger attacks. These findings suggest that coupling classification, localization, and post-processing disruption is necessary for achieving robust and stable attack performance.

\subsection{Additional Study} \label{sec:additional_studies}

In this section, we evaluate the robustness and reliability of our proposed method by examining its performance consistency across multiple random initializations. The evaluation is conducted on the INRIA dataset using the YOLOv5 detector, with four different random seeds to ensure statistical significance and eliminate potential bias from specific initialization conditions. This comprehensive analysis addresses a critical aspect of adversarial attack evaluation, as the effectiveness of patch-based attacks can sometimes be sensitive to the initial parameter settings and optimization trajectories.

\begin{figure}[t]
    \centering
    \includegraphics[scale=0.28]{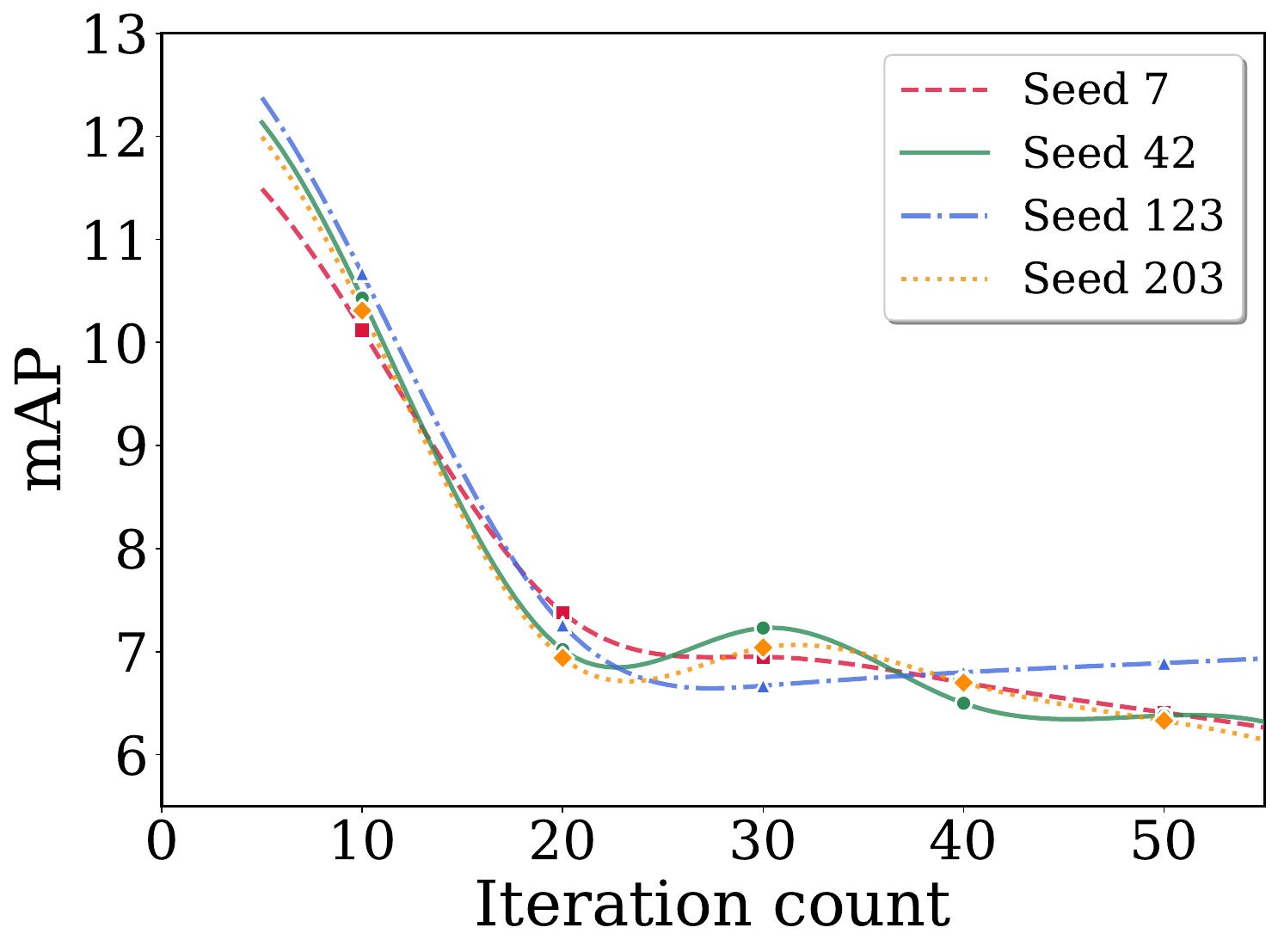}
    \caption{Performance consistency across multiple random seeds of our method.
    }
    \label{Fig:seeds}
      \vspace{-0.4cm}
\end{figure}

\Cref{Fig:seeds} illustrates the performance consistency of our method across different batch sizes with multiple random seeds (42, 7, 123, and 203). These seeds are carefully selected to represent diverse initialization scenarios and provide a comprehensive assessment of our method's stability. The results demonstrate that our method maintains remarkably stable performance across different random initializations, with all four seeds showing similar convergence trends and comparable mAP values throughout the training process. The convergence curves exhibit consistent patterns regardless of the initial random state, indicating that our triple-loss optimization framework successfully guides the patch generation process toward effective adversarial patterns.

%% file: Section/5-conclusion.tex
\section{Limitations and Conclusions}
\label{sec:Conclusion}

In this paper, we propose a novel adversarial patch generation method targeting pedestrian detection systems in the physical world. 
By jointly attacking multiple stages of the object detector’s decision-making pipeline, the proposed approach effectively achieves pedestrian invisibility. 
We introduce a triple-cooperative loss mechanism, consisting of confidence suppression, bounding box perturbation amplification, and Non-Maximum Suppression  disruption. 
These components jointly degrade the detector’s performance across classification, localization, and post-processing stages.
Furthermore, we incorporate spatial-domain regularization constraints, combining appearance consistency loss to suppress high-frequency noise and blend the adversarial patch seamlessly with natural clothing textures, thereby enhancing physical robustness.
Experimental results under various mainstream detection models and challenging physical environments demonstrate that the proposed method significantly reduces the mAP and maintains a high attack success rate across diverse scenarios.
However, this approach still has certain limitations, as the naturalness of the adversarial patch can be further improved. Future work may explore disguising the patch as common patterns to enhance its visual plausibility and practical applicability.

